\documentclass[AMA,STIX1COL,doublespace]{WileyNJD-v2}

\usepackage{longtable}

\usepackage{placeins}
\newcommand{\ie}{\textit{i.e., }}
\newcommand{\eg}{\textit{e.g., }}

\articletype{Research article}

\received{2020-xx-xx}

\revised{2020-xx-xx}

\accepted{2020-xx-xx}

\begin{document}

\title{When to Impute? Imputation before and during cross-validation}

\author[a]{Byron C. Jaeger*}
\author[b]{Nicholas J. Tierney}
\author[c]{Noah R. Simon}

\authormark{Jaeger \emph{et al}.}

\address[a]{Department of Biostatistics, University of Alabama at Birmingham,
Birmingham, Alabama}
\address[b]{Department of Econometrics and Business Statistics, Monash University,
Melbourne, Victoria, Australia. ARC Centre of Excellence for
Mathematical and Statistical Frontiers (ACEMS), Brisbane, Queensland,
Australia}
\address[c]{Department of Biostatistics, University of Washington, Seattle,
Washington}

\corres{*Byron C. Jaeger. \email{bcjaeger@uab.edu}}

\presentaddress{327M Ryals Public Health Building 1665 University Blvd Birmingham,
Alabama 35294-0022}

\abstract{Cross-validation (CV) is a technique used to estimate generalization
error for prediction models. For pipeline modeling algorithms
(\ie modeling procedures with multiple steps), it has been recommended
the \emph{entire} sequence of steps be carried out during each replicate
of CV to mimic the application of the entire pipeline to an external
testing set. While theoretically sound, following this recommendation
can lead to high computational costs when a pipeline modeling algorithm
includes computationally expensive operations, \eg imputation of missing
values. There is a general belief that unsupervised variable selection
(\ie ignoring the outcome) can be applied before conducting CV without
incurring bias, but there is less consensus for unsupervised imputation
of missing values. We empirically assessed whether conducting
unsupervised imputation prior to CV would result in biased estimates of
generalization error or result in poorly selected tuning parameters and
thus degrade the external performance of downstream models. Results show
that despite optimistic bias, the reduced variance of imputation before
CV compared to imputation during each replicate of CV leads to a lower
overall root mean squared error for estimation of the true external
\(R^2\) and the performance of models tuned using CV with imputation
before versus during each replication is minimally different. In
conclusion, unsupervised imputation before CV appears valid in certain
settings and may be a helpful strategy that enables analysts to use more
flexible imputation techniques without incurring high computational
costs.}


\maketitle

\section{Introduction}

In evaluating the performance of a predictive modeling algorithm, it is
understood that so-called training error (the predictive error measured
on observations used to fit the model) is a poor proxy for
generalization error (the performance of the model on future,
as-yet-unseen, observations).\citep{kuhn2013applied} The training error
of a model will often be overly optimistic for the generalization error.
As such, it is standard practice to use sample-splitting methods to
estimate generalization error. These methods train and test models using
separate datasets. \(v\)-fold Cross-validation (CV) is a common
sample-splitting method that partitions a dataset into \(v\)
non-overlapping subsets (\textit{i.e., }folds).\cite{arlot2010survey}
Each fold is then used as an internal assessment set for a modeling
algorithm developed using data from the \(v-1\) remaining folds.
Aggregating errors from all \(v\) replications of this procedure
provides an estimate of the modeling algorithm's generalization error,
making \(v\)-fold CV an effective technique to `tune' modeling
algorithms (\textit{i.e., }select optimal values for parameters that
govern the algorithm's fitting procedure).

Machine learning analyses often involve `pipeline' modeling algorithms:
multi-step modeling procedures that may include data pre-processing,
predictor variable selection and/or transformation, model fitting, and
ensembling.\citep{mlr3} For example, a pipeline may begin by centering
and scaling predictor values, then filter out redundant correlated
predictors, and finally fit a regression model to the remaining data. To
estimate the generalization error of pipeline modeling algorithms using
CV, it is recommended that the \emph{entire sequence} of steps be
carried out during each replicate of CV to mimic the application of the
entire pipeline to an external testing set. However, it has been
suggested that unsupervised variable selection steps
(\textit{i.e., }steps that ignore the outcome variable) can be applied
before conducting CV without incurring bias.\citep{hastie2009elements}
Since unsupervised predictor variable selection does not involve outcome
variables, it does not give the selected predictors an unfair advantage
during CV.

Missing data (MD) occur frequently in machine learning analyses, and
several learning algorithms (e.g., regression) are incompatible with MD.
Imputation is a technique that replaces MD with estimated values, and is
often among the most computationally expensive operations in pipeline
modeling algorithms. For example, the \texttt{missForest} imputation
algorithm may fit one random forest model for each column that contains
MD.\citep{stekhoven2011missforest} Computational expense of applying
\texttt{missForest} or other complex imputation strategies during each
replicate of CV scales poorly and may lead analysts to prefer more
convenient but less effective strategies to handle MD. A more
computationally efficient approach would be to implement `unsupervised
imputation' (\textit{i.e., }imputing MD without accessing outcome
information) \emph{before} conducting CV. Ordering operations this way
could result in substantially faster training and tuning costs for
pipeline modeling algorithms because imputation is only performed once
rather than once per CV fold. However, it is unclear whether an
unsupervised operation that modifies the training data
(\textit{i.e., }imputation) rather than removing columns from the
training data (\textit{i.e., }variable selection) can be applied prior
to CV without causing model error estimates to become overly optimistic.

The aim of this paper is to assess whether unsupervised imputation
before versus during CV causes bias in estimation of a modeling
pipeline's generalization error. We also investigate whether
unsupervised imputation before versus during CV can result in poorly
selected tuning parameters and thus degrade the external performance of
downstream models. To achieve these aims, we conduct empirical studies
of simulated and real data assessing whether unbiased generalization
error estimates are obtained if unsupervised imputation is implemented
before CV, a strategy we will refer to as
$\texttt{I}\!\!\rightarrow\!\texttt{CV}$. We compare estimated pipeline
error according to $\texttt{I}\!\!\rightarrow\!\texttt{CV}$\space with
estimated pipeline error when unsupervised imputation is applied
\emph{during each replicate} of CV, a strategy we will refer to as
$\texttt{CV}\!\circlearrowright\!\texttt{I}$. All scripts and data
involved in the current analysis are publicly available and all results
are reproducible (See
\href{https://github.com/bcjaeger/Imputation-and-CV}{first author's
GitHub}). Our analysis also introduces and applies the \texttt{ipa} R
package (\textbf{i}mputation for \textbf{p}redictive
\textbf{a}nalytics), which provides functions to create single or
multiple imputed training and testing sets for prediction modeling.

The rest of this manuscript is organized as follows. In Section
\ref{sec:missing_data}, we discuss MD mechanisms and prevailing MD
strategies for statistical inference and machine learning. In Section
\ref{sec:oop}, we explicitly map the order of operations for
$\texttt{CV}\!\circlearrowright\!\texttt{I}$\space and
$\texttt{I}\!\!\rightarrow\!\texttt{CV}$. In section \ref{sec:sim}, we
conduct a simulation study to assess empirical differences between
$\texttt{CV}\!\circlearrowright\!\texttt{I}$\space and
$\texttt{I}\!\!\rightarrow\!\texttt{CV}$. The two procedures are
compared using real data in Section \ref{sec:app}. Last, in Section
\ref{sec:discuss}, we organize the data from preceding sections to form
recommendations for practitioners.

\section{Missing data} \label{sec:missing_data}

\paragraph{Missing data mechanisms}

MD mechanisms were first formalized by Rubin,\citep{rubin1976inference}
who developed a framework to analyze MD that supposes each data point
has some probability of being missing. If the probability of missingness
is unrelated to the data (\textit{i.e., }all data are equally likely to
be missing), then the data are missing completely at random (MCAR). When
the probability of missingness is related to observed variables in the
data (\textit{i.e., }all data within observed groups are equally likely
to be missing), the data are missing at random (MAR). If the probability
of missingness is determined by reasons that are unknown or unobserved,
the data are missing not at random (MNAR). To illustrate, if a patient's
data are missing because of clerical data entry error, the patient's
data are MCAR. If instead a doctor chose not to measure the patient's
labs because the patient was too young, the patient's data are MAR. If
the patient missed the appointment because the patient was too sick, the
patient's data are MNAR. In the context of statistical learning,
previous findings have shown that when data are MNAR, imputation alone
is often less effective than incorporating features that characterize
missing patterns (\textit{e.g., }missingness incorporated as an
attribute).\citep{twala2008good, twala2009empirical, tang2017random}
Since the primary aim of the current study is to assess the differences
between two implementations of imputation
(\textit{i.e., }$\texttt{I}\!\!\rightarrow\!\texttt{CV}$\space and
$\texttt{CV}\!\circlearrowright\!\texttt{I}$), we focus analyses on
cases where data are MAR or MCAR.

\paragraph{Missing data strategies for statistical inference}

The primary objective for statistical inference in the presence of MD is
to obtain valid test statistics for statistical hypotheses. Imputation
to the mean and, more broadly, MD strategies that create a single
imputed value, have been shown to increase type I errors
(\textit{i.e., }rejecting a true null hypothesis) for inferential
statistics by artificially reducing the variance of observed data and
ignoring the uncertainty attributed to MD, respectively. Multiple
imputation, a widely recommended strategy to handle MD for statistical
inference, is capable of producing valid test statistics when data are
MCAR or MAR because it can simultaneously address these two
shortcomings. The `accuracy' of imputed values is not critical for the
success of multiple imputation, given sufficient estimates of
conditional distributions \citep{van2018flexible}. Instead, the
consistency of the estimated covariance matrix for regression
coefficients makes this strategy ideal for statistical inference.

\paragraph{Missing data strategies for statistical prediction}

The primary objective for statistical prediction in the presence of MD
is to develop a prediction function that accurately generalizes to
external data, which may or may not contain missing values (see Section
\ref{subsec:testing_data}). In contrast to statistical inference, single
imputation is often used for prediction models.\citep{kuhn2019feature}
Moreover, imputation strategies with greater accuracy often lead to
better performance of downstream models (\textit{i.e., }models fitted to
the imputed data). For example, Jerez et al.~found single imputation
using machine learning models provided superior downstream model
prognostic accuracy compared to multiple imputation based on regression
and expectation maximization.\citep{jerez2010missing} The results of
this analysis exemplify a perspective that will be taken throughout the
current study. Namely, the authors treated imputation strategies as
components of the modeling pipeline with parameters that can be `tuned'
in the same manner as a prediction model.

\section{Order of Operations} \label{sec:oop}

\begin{figure}
\includegraphics[width=1\linewidth]{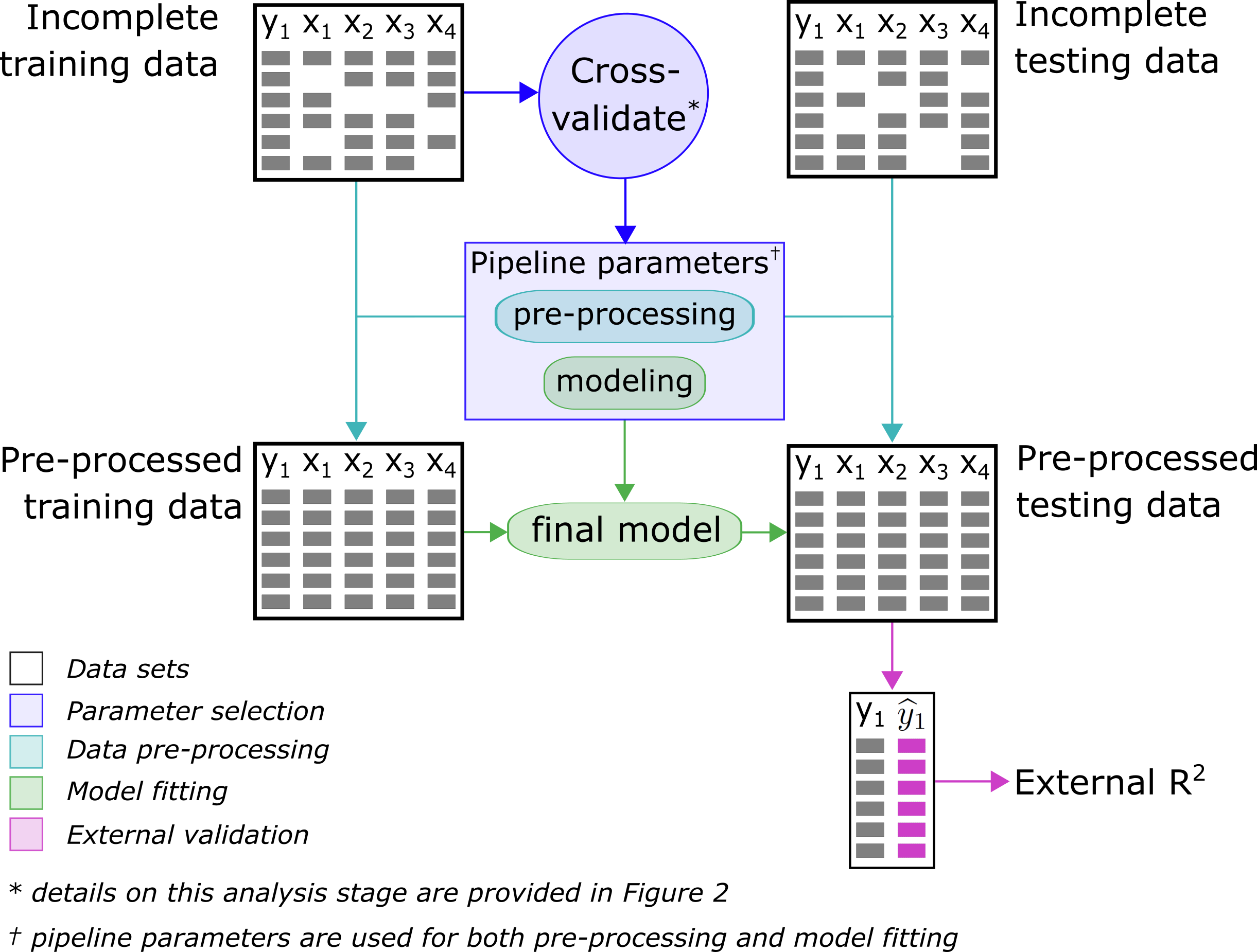} 
\caption{A workflow to develop and validate a pipeline modeling algorithm. Pipeline parameter values may be set apriori or determined empirically using cross validation. Once parameter values are fixed, a final model is developed by training the modeling pipeline using the training data. The final model is externally validated by assessing the accuracy of its predictions in the testing data.}
\label{fig:workflow_ml}
\end{figure}

In the context of statistical prediction, analysts usually work with a
training set and an external testing set. A pipeline modeling algorithm
developed with data from the training set can be externally validated
using data from the testing set. Workflows to develop and validate a
pipeline model may include three steps: (1) selection of pipeline
parameter values (\textit{i.e., }parameters relevant to any operation in
the pipeline, including data pre-processing), (2) developing a final
model by training the modeling pipeline using the training data, and (3)
externally validating the final model by assessing the accuracy of its
predictions with the testing data. This workflow is described in
\textbf{Figure} \ref{fig:workflow_ml}.

Pipeline parameter values may be set apriori or determined empirically
(\textit{i.e., }tuned) using resampling, \textit{e.g., }by leveraging
\(v\)-fold CV. We refer to the \(v-1\) folds and 1 remaining fold used
to internally train and test a modeling algorithm as \textbf{analysis}
and \textbf{assessment} sets, respectively, to avoid notation abuse of
the terms ``training'' and ``testing.'' \cite{breiman} If CV is applied
to facilitate selection of pipeline parameter values, it is critical
that analysis data are separated from assessment data before any
`learning' is done. The entire \emph{supervised} pipeline must be run
using only the assessment data. This applies both to supervised data
pre-processing steps (\textit{e.g., }selecting all variables with high
correlation to the outcome) as well as supervised modeling procedures
(\textit{e.g., }regression). `Data leakage' can occur when outcome
information from the assessment set is leveraged to modify the analysis
set, \textit{e.g., }supervised variable selection is performed on a
stacked set comprising analysis and assessment data, rather than just
the assessment data.\cite{hastie2009elements} There are a number of
examples showing wildly optimistic estimates of generalization error
because of data leakage.\cite{neunhoeffer2019cross} In scenarios with a
larger number of features, even simple methodologies such as selecting
those features with high individual correlation to the outcome can
induce substantial overoptimism.

\begin{figure}
\centering
\includegraphics[width=0.85\linewidth]{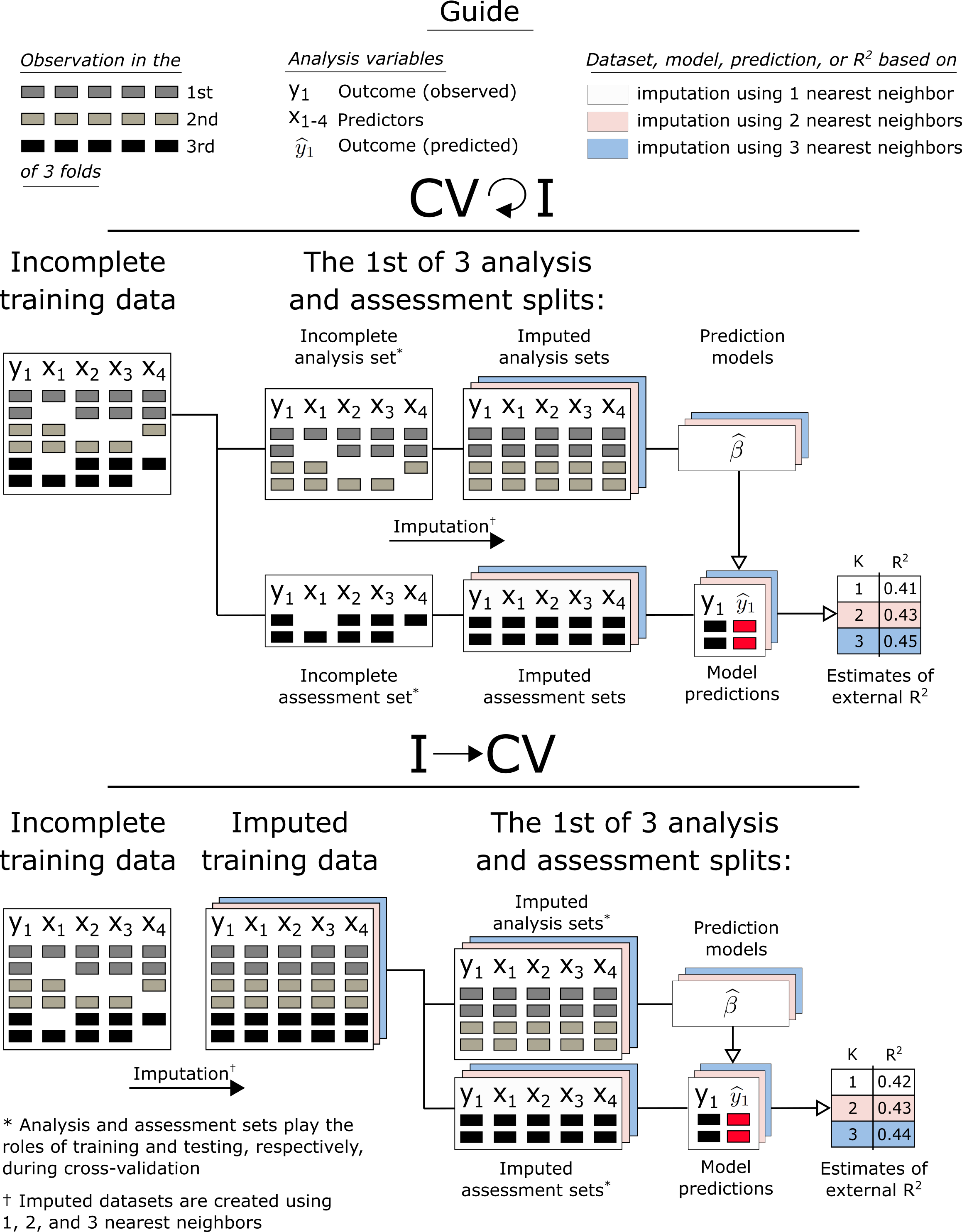} 
\caption{Workflows for cross validation (CV) incorporating imputation of missing values. The difference in the workflows is where imputation is performed. The standard workflow, $\texttt{CV}\!\circlearrowright\!\texttt{I}$, imputes missing values during each replicate of CV. The experimental workflow, $\texttt{I}\!\!\rightarrow\!\texttt{CV}$, imputes missing values prior to CV. Critically, $\texttt{I}\!\!\rightarrow\!\texttt{CV}$\space means imputation happens once, whereas in $\texttt{CV}\!\circlearrowright\!\texttt{I}$\space the imputation procedure occurs for each fold, adding computational time.}
\label{fig:workflow_cv_bothways}
\end{figure}

To remove any possibility of data leakage, all steps of the pipeline may
be performed in analysis and assessment sets, separately, within each
replicate of CV. For example, consider centering and scaling predictor
variables such that they have zero mean and unit variance. As these
operations do not involve the outcome, they are entirely unsupervised.
Nevertheless, centering and scaling operations are usually completed in
analysis and assessment sets, separately, during each replicate of CV.
Specifically, the means and standard deviations are computed using the
analysis data and then those values are applied to center and scale
predictors in both the analysis and assessment sets. We refer to this
traditional implementation of CV as
$\texttt{CV}\!\circlearrowright\!\texttt{I}$\space and refer to the
experimental implementation of CV (\textit{i.e., }one where unsupervised
imputation occurs before CV begins) as
$\texttt{I}\!\!\rightarrow\!\texttt{CV}$\space (\textbf{Figure}
\ref{fig:workflow_cv_bothways}). Regardless of which implementation is
applied, the output of CV is a set of pipeline parameter values and an
estimate of generalization error. The pipeline parameter values are
subsequently used to develop and validate a final prediction model using
the full training set and testing set, respectively. Differences in
parameter values selected by competing CV strategies
(\textit{i.e., }$\texttt{CV}\!\circlearrowright\!\texttt{I}$\space or
$\texttt{I}\!\!\rightarrow\!\texttt{CV}$) may have measurable impact on
the generalization error of their downstream models.

\subsection{Testing data} \label{subsec:testing_data}

Ideally, external testing data will not contain MD, and imputation will
not be necessary. However, If MD are present in the external testing
data, additional steps may be taken to engage with them. One may impute
missing values in the testing data using (1) only the training data, (2)
only the testing data, or (3) using both training and testing data. It
is common to use only the training data to impute missing values in the
testing data. However, some imputation procedures can't be discretely
separated into two steps, one that develops an imputation model and
another that applies it to new data. For example, matrix decomposition
methods such as \texttt{softImpute} merely take a matrix with missing
entries and fill them in.\citep{softImpute} To apply these types of
imputation procedures, approach (2) or (3) may be taken. Notably,
$\texttt{CV}\!\circlearrowright\!\texttt{I}$\space uses only the
analysis data to impute missing values in the assessment data
(\textit{i.e., }approach 1), and
$\texttt{I}\!\!\rightarrow\!\texttt{CV}$\space uses a stacked version of
the analysis and assessment data (\textit{i.e., }all of the training
data; approach 3) to impute missing values. Following CV, our analyses
strictly implement approach 1 to engage with missing values in the
testing data.

\section{Simulated experiments} \label{sec:sim}

The goal of the current simulation study was to assess empirical
differences between
$\texttt{CV}\!\circlearrowright\!\texttt{I}$\space and
$\texttt{I}\!\!\rightarrow\!\texttt{CV}$\space following a published
protocol.\cite{morris2019using} Our primary objective was to measure and
compare how well each strategy (1) approximated a model's generalization
error and (2) selected parameters (both for imputation and modeling)
that would maximize downstream model accuracy. To complete item (1), we
assessed estimation of the external \(R^2\). We used bias, variance, and
root-mean-squared error (RMSE) to quantify estimation accuracy. The RMSE
provides an overall assessment of estimation accuracy that depends on
both bias and variance. To complete item (2), we compared the
performance (\textit{i.e., }the external \(R^2\)) of downstream models
whose tuning parameters were selected using
$\texttt{CV}\!\circlearrowright\!\texttt{I}$\space versus
$\texttt{I}\!\!\rightarrow\!\texttt{CV}$.

\subsection{Data-generating mechanisms} \label{subsec:data_gen}

Consider the linear regression model, where a continuous outcome vector
\(\textbf{y} = \lbrace y_1, y_2, \ldots, y_N\rbrace\) is generated by a
linear combination of predictor variables
\(\textbf{X} = \left[ \textbf{x}_1, \textbf{x}_2, \ldots \textbf{x}_p \right]\).
This functional relationship is often expressed as
\[\textbf{Y} = \textbf{X} \beta + \varepsilon,\] where \(\beta\) is a
\(p \times 1\) vector of regression coefficients and \(\varepsilon\) is
a normally distributed \(N \times 1\) zero-mean random error vector. In
practice, \(\textbf{X}\) often has some `junk' variables that are not
related to the outcome. To make our simulations similar to applied
settings, we generated normally distributed variables that had no
relation to the simulated outcome. We fixed the number of true predictor
variables at 10, the standard error of \(\varepsilon\) at 1, and set
\(\beta = [-1.00, -0.78, -0.56, -0.33, -0.11, 0.11, 0.33, 0.56, 0.78, 1.00]\)
throughout the simulation study. Columns of \(\textbf{X}\) were
generated from a multivariate normal distribution with a first order
autoregressive correlation structure. Specifically, the correlation
between columns \(\textbf{x}_i\) and \(\textbf{x}_j\) was
\(\rho^{\left| i-j \right|}\), where \(\rho\) was set to 3/4 throughout
the study. We applied this design to generate a training set of varying
size (100, 500, 1000, or 5000) along with an external validation set
comprising 10,000 observations in each simulated replicate.

\paragraph{Data generation scenarios}

We created three data-generation `scenarios'. In scenario 1, the
observed data are independent and identically distributed (iid). In
scenario 2, the data are iid conditional on an observed grouping
variable. A total of 11 groups are formed, one in the validation set and
the remaining 10 in the training set. Each group is characterized by a
randomly generated mean value for its predictor variables. During CV,
the observed groups are separated into ten folds to mimic the prediction
of outcomes in a population with different characteristics. Scenario 3
is identical to scenario 2 except that the grouping variable is latent.
Consequently, CV does not break the observed groups into separate folds
for scenario 3.

\paragraph{Amputing data}

We applied the \texttt{ampute} function from the \texttt{mice} R package
to generate missing values in simulated data.\cite{mice} In each
replicate, 90\% of observations comprised at least one missing value. We
designated up to \(p\) MD patterns randomly in each simulation
replicate, where \(p\) is the number of non-outcome columns in the
simulated data. A MD pattern indicates which of the \(p\) predictor
variables are set to missing. For each MD pattern, the number of missing
variables was randomly set to an integer ranging from 1 to \(p/2\). This
procedure usually induced missing values in 30-50\% of the data. When
data were MAR, we applied the default method for the \texttt{ampute}
function (\texttt{ampute.default.weights}) to induce missingness based
on the observed variables. Throughout the experiment, we applied the
same missing patterns and MD mechanism in the training set and the
external validation set.

\paragraph{Modeling procedure}

We applied \(k\)-nearest-neighbor imputation to handle MD and least
absolute shrinkage and selection operator (LASSO) regression to develop
prediction functions throughout the simulated
experiments.\cite{tibshirani1996regression} The LASSO model is an
appropriate model for these simulations since all data were generated
with linear, independent effects. Nearest neighbor aggregation based on
Gower's distance was used to form imputed values in the training and
testing set and also in the analysis and assessment set for
$\texttt{CV}\!\circlearrowright\!\texttt{I}$.\cite{gower1971general} We
created one imputed dataset for each
\(k \in \lbrace 1, 2, \ldots, 35\rbrace\). We selected a value for the
regularization parameter \(\lambda\) in each imputed dataset,
separately, using 10-fold CV
(\textit{i.e., }\texttt{cv.glmnet}).\cite{glmnet} The \(\lambda\) value
selected was the one that minimized the model's cross-validated RMSE.
The value of \(k\) that minimized cross-validated RMSE was used to
impute the entire training set prior to fitting a final
\texttt{cv.glmnet} model.

\paragraph{Analysis plan}

We varied the scenario (1, 2, or 3; described above), missing mechanism
(MCAR or MAR), ratio of predictor variables to junk variables (1:1, 1:4,
and 1:49; junk variables have no relationship to the simulated outcome),
and the number of training observations (\(N\) = 100, 500, 1,000,
5,000). We present results for each of 72 settings determined by these
parameters and also provide overall summary statistics for scenarios 1,
2, and 3 when data are MCAR and MAR (\textit{i.e., }aggregating over
training sample size and predictor to noise ratio). In each simulation
replication, we computed the true external \(R^2\) in the validation set
for each potential value of nearest neighbors
(\textit{i.e., }\(k \in \lbrace 1, 2, \ldots, 35 \rbrace\)). We also
estimated external \(R^2\) for each value of \(k\) using
$\texttt{CV}\!\circlearrowright\!\texttt{I}$\space and
$\texttt{I}\!\!\rightarrow\!\texttt{CV}$, separately, to evaluate how
well these CV procedures estimated the true external \(R^2\). We
assessed the difference between estimated external \(R^2\) according to
$\texttt{CV}\!\circlearrowright\!\texttt{I}$\space and
$\texttt{I}\!\!\rightarrow\!\texttt{CV}$\space as well as the bias,
variance, and root-mean-squared error (RMSE) of these estimates. Last,
we investigated the accuracy of downstream models when
$\texttt{CV}\!\circlearrowright\!\texttt{I}$\space and
$\texttt{I}\!\!\rightarrow\!\texttt{CV}$\space were applied to select
the number of neighbors to use for imputation and then the
regularization parameter for a penalized regression model.

\subsection{Results} \label{subsec:sim_results}

Overall, a total of 143,253 out of 144,000 (99.5\%) simulation
replicates were completed over a span of 52,979 computing hours.
Incomplete replicates were not analyzed, as these were replicates where
at least one of the amputation, imputation, or prediction models did not
converge. Across all replicates, the median
(25\(^{\text{th}}\)-75\(^{\text{th}}\) percentile) number of seconds
used to form imputed data using
$\texttt{CV}\!\circlearrowright\!\texttt{I}$\space and
$\texttt{I}\!\!\rightarrow\!\texttt{CV}$\space were 27 (11 - 225) and
2.3 (0.69 - 22), respectively, a ratio of 11 (9.8 - 14). Using the full
imputed training set, the median (25\(^{\text{th}}\)-75\(^{\text{th}}\)
percentile) number of seconds needed to tune \texttt{glmnet} models
using CV and fit a final model to the full training set was 4.8 (3.3 -
55), verifying our earlier claim that complex imputation procedures
often require more time than modeling procedures.

\begin{figure}
\includegraphics[width=1\linewidth]{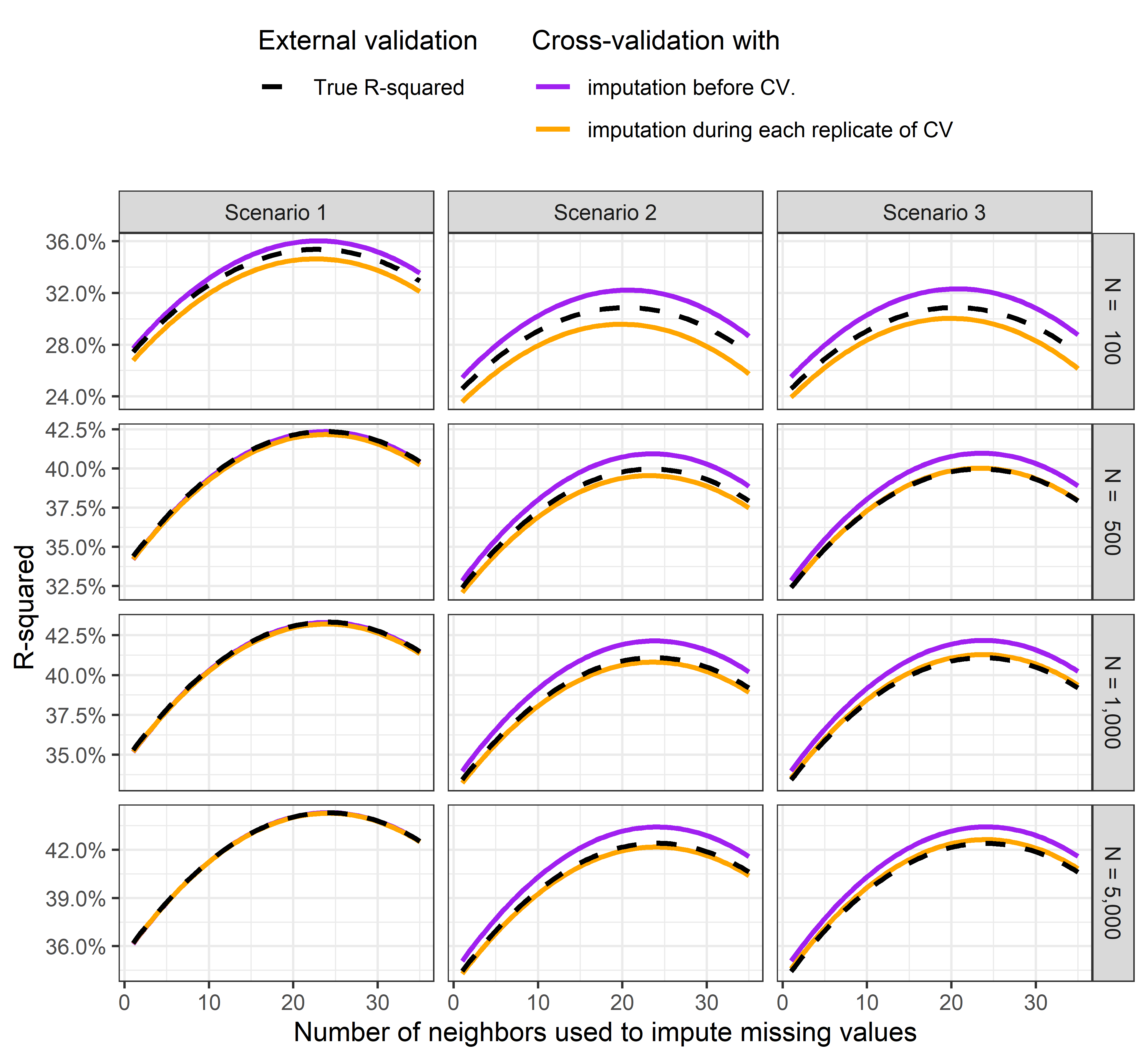} 
\caption{External generalization error and internal estimates of generalization error using $\texttt{I}\!\!\rightarrow\!\texttt{CV}$\space and $\texttt{CV}\!\circlearrowright\!\texttt{I}$. The $R^2$ values are plotted as a function of the number of nearest neighbors used to impute missing data, and the panel rows show results with 100, 500, 1000, and 5000 observations in the training data. The scenarios are described in Section \ref{subsec:data_gen}. The peaks of all three curves consistently appear to be at the same number of neighbors. While $\texttt{I}\!\!\rightarrow\!\texttt{CV}$\space error estimates have a slight positive bias, as noted in section \ref{subsec:sim_results}, they also have less variability than error estimates using $\texttt{CV}\!\circlearrowright\!\texttt{I}$.}
\label{fig:sim_r2}
\end{figure}

Across all scenarios, the mean external \(R^2\) ranged from 0.233 to
0.443 (\textbf{Table} \ref{tab:ext_rsq}). External \(R^2\) values
increased with larger training set size and higher ratio of predictor
variables to junk variables. Notably, the mean external \(R^2\) values
in scenario 1 were uniformly greater than corresponding mean external
\(R^2\) values in scenarios 2 and 3, and the maximum difference between
mean external \(R^2\) values in scenario 2 versus scenario 3 was 0.0003.
The mean absolute difference between external \(R^2\) estimates using
$\texttt{CV}\!\circlearrowright\!\texttt{I}$\space and
$\texttt{I}\!\!\rightarrow\!\texttt{CV}$\space shrunk towards zero as
the size of the training set increased (\textbf{Table}
\ref{tab:cv_diffs}). The differences between
$\texttt{CV}\!\circlearrowright\!\texttt{I}$\space and
$\texttt{I}\!\!\rightarrow\!\texttt{CV}$\space were lowest in scenario 1
and greatest in scenario 2. These patterns were also present in visual
depictions of external \(R^2\) portrayed as a function of \(k\)
neighbors (\textbf{Figure} \ref{fig:sim_r2}).

\paragraph{Bias, variance, and RMSE}

For scenario 1, the overall bias of \(R^2\) estimates under MCAR using
$\texttt{CV}\!\circlearrowright\!\texttt{I}$\space was -0.00136 versus
0.00233 using
$\texttt{I}\!\!\rightarrow\!\texttt{CV}$\space (\textbf{Table}
\ref{tab:bias}). When the data were MAR, the overall biases were
-0.00080 for $\texttt{CV}\!\circlearrowright\!\texttt{I}$\space versus
0.00300 for $\texttt{I}\!\!\rightarrow\!\texttt{CV}$\space. In scenarios
2 and 3, the bias of
$\texttt{CV}\!\circlearrowright\!\texttt{I}$\space was lower than that
of $\texttt{I}\!\!\rightarrow\!\texttt{CV}$\space, and
$\texttt{I}\!\!\rightarrow\!\texttt{CV}$\space consistently provided
overly optimistic error estimates. The overall standard deviation of
\(R^2\) estimates was higher for
$\texttt{CV}\!\circlearrowright\!\texttt{I}$\space versus
$\texttt{I}\!\!\rightarrow\!\texttt{CV}$\space in all three scenarios
and both missing data mechanisms ( \textbf{Table} \ref{tab:variance}).
The difference in standard deviation was most pronounced in scenario 3
when data were MCAR (0.07314
{[}$\texttt{CV}\!\circlearrowright\!\texttt{I}${]} versus 0.06747
{[}$\texttt{I}\!\!\rightarrow\!\texttt{CV}${]}). Despite the optimistic
bias of $\texttt{I}\!\!\rightarrow\!\texttt{CV}$\space in scenario 2,
the reduced variance of this approach lead to a lower overall RMSE for
external \(R^2\) compared to
$\texttt{CV}\!\circlearrowright\!\texttt{I}$\space (\textbf{Table}
\ref{tab:rmse}). When the data were MCAR in scenario 2,
$\texttt{CV}\!\circlearrowright\!\texttt{I}$\space and
$\texttt{I}\!\!\rightarrow\!\texttt{CV}$\space obtained RMSEs of 0.05738
and 0.05593, respectively. Similarly, when the data were MAR in scenario
2, overall RMSE values were 0.05751 and 0.05572.

\paragraph{Downstream model performance}

When $\texttt{CV}\!\circlearrowright\!\texttt{I}$\space and
$\texttt{I}\!\!\rightarrow\!\texttt{CV}$\space were applied to select
tuning parameters, the overall mean external \(R^2\) was higher using
$\texttt{CV}\!\circlearrowright\!\texttt{I}$\space in 22 out of 78
comparisons (28\%; \textbf{Table} \ref{tab:tune}). However, the
differences in mean external \(R^2\) between models tuned using
$\texttt{CV}\!\circlearrowright\!\texttt{I}$\space and
$\texttt{I}\!\!\rightarrow\!\texttt{CV}$\space were relatively minor.
For instance, the greatest overall difference in mean \(R^2\) between
downstream models occurred in scenario 1 when the data were MAR
(absolute difference in model \(R^2\): 0.00028; relative difference in
model \(R^2\) : 0.07\%).

\section{Real data experiments} \label{sec:app}

The goal of the current resampling study was to repeat comparisons
summarized in Section \ref{sec:sim} between
$\texttt{CV}\!\circlearrowright\!\texttt{I}$\space and
$\texttt{I}\!\!\rightarrow\!\texttt{CV}$\space using real, publicly
accessible data. A secondary objective was to assess how much results
would change if different modeling strategies were applied.

\paragraph{Ames, Iowa housing data}

The data we use in this resampling study describe the sale of individual
residential property in Ames, Iowa from 2006 to 2010. The entire set
contains 2930 observations and 80 variables (23 nominal, 23 ordinal, 14
discrete, and 20 continuous) that can be leveraged to predict the sale
price of homes.\cite{de2011ames} We used a cleaned version of the Ames
data for our own analyses by applying the \texttt{make\_ames()}
function, available in the \texttt{AmesHousing} R
package.\cite{AmesHousing} We also log-transformed the skewed sale price
outcome.

\paragraph{Analysis plan}

We conducted a resampling study where the Ames housing data was randomly
split into training (\(N = 2198\), 75\%) and testing (\(N = 732\), 25\%)
sets in each of 5,000 iterations. In each resampling replicate, we
implemented two separate modeling strategies to develop prediction
functions using the training set: (1) un-penalized linear regression and
(2) random forests.\cite{breiman2001random} We also implemented two
imputation strategies: (1) nearest neighbor imputation using 1, 2,
\ldots, 35 neighbors and (2) mean and mode imputation for numeric and
nominal variables, respectively. In addition to imputation, data were
pre-processed by lumping values in discrete variables into an `other'
category if the value accounted for less than 10\% of the observed
values. Both CV techniques
(\textit{i.e., }$\texttt{CV}\!\circlearrowright\!\texttt{I}$\space and
$\texttt{I}\!\!\rightarrow\!\texttt{CV}$) were implemented to estimate
the external generalization error of the linear regression and random
forest models when nearest neighbor imputation was applied.

\paragraph{Amputing data}

The training and testing data were amputed in the same manner using four
prototypes of missingness. The prototypes were characterized by having
missing values for all variables describing the house (1) lot and
garage, (2) longitude and latitude, (3) basement and year built, and (4)
overall quality and general above ground square footage. We restricted
our prediction models to consider only the 30 predictor variables
belonging to at least one of these missing prototypes.

\subsection{Results}

A total of 5000 out of 5000 (100\%) resampling replicates were completed
over a span of 927 computing hours. Across all replicates, the mean
number of minutes used to form imputed data using
$\texttt{CV}\!\circlearrowright\!\texttt{I}$\space and
$\texttt{I}\!\!\rightarrow\!\texttt{CV}$\space were 10 and 1.0,
respectively. The mean (standard deviation) external \(R^2\) for the
linear regression and random forest models were 0.7641 (0.0265) and
0.8088 (0.0205), respectively. Overall, both CV techniques slightly over
and under estimated the external \(R^2\) value when linear regression
and random forests were applied, respectively. For linear regression,
the mean (standard deviation) bias was -0.0015 (0.0343) and -0.0021
(0.0343) for $\texttt{CV}\!\circlearrowright\!\texttt{I}$\space and
$\texttt{I}\!\!\rightarrow\!\texttt{CV}$, respectively. The standard
deviations of error estimates were 0.0115 and 0.0115, respectively. For
random forests, the mean (standard deviation) bias was 0.0009 (0.0255)
and 0.0005 (0.0255) for
$\texttt{CV}\!\circlearrowright\!\texttt{I}$\space and
$\texttt{I}\!\!\rightarrow\!\texttt{CV}$, respectively. The standard
deviation of error estimates were 0.0084 and 0.0086, respectively.

\begin{figure}
\includegraphics{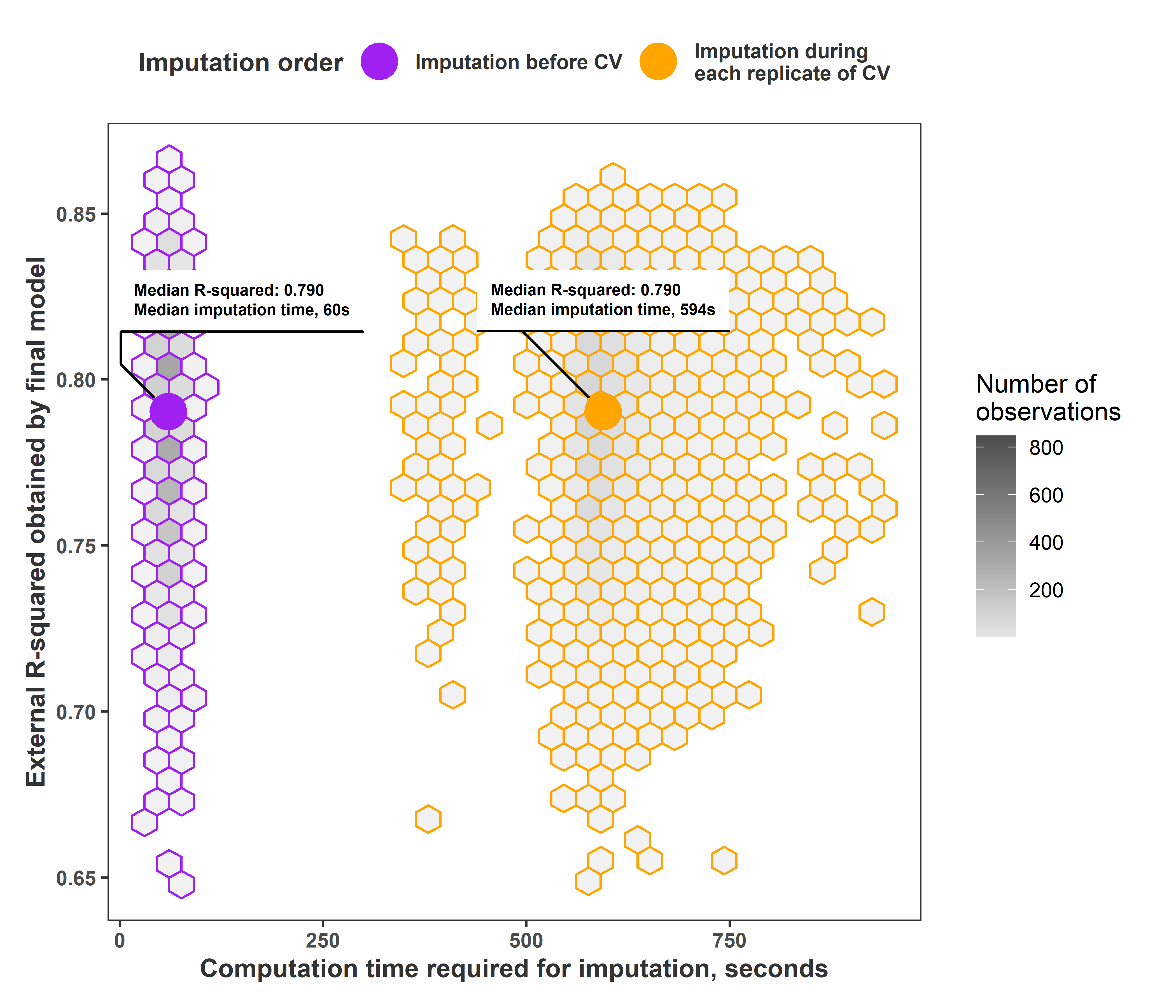} 
\caption{External generalization error (y-axis) and time required to impute missing values (x-axis) for the Ames housing data. $\texttt{I}\!\!\rightarrow\!\texttt{CV}$\space and $\texttt{CV}\!\circlearrowright\!\texttt{I}$\space were applied, separately, to select $k$, the number of nearest neighbors used to impute missing values, prior to fitting and externally validating a prediction model. While the median generalization error is practically equivalent regardless of which CV method was used, the time required for imputation is approximately 10 (\textit{i.e., }the number of folds in CV) times higher using $\texttt{CV}\!\circlearrowright\!\texttt{I}$\space versus $\texttt{I}\!\!\rightarrow\!\texttt{CV}$.}
\label{fig:ames_cmp_time}
\end{figure}

When $\texttt{CV}\!\circlearrowright\!\texttt{I}$\space and
$\texttt{I}\!\!\rightarrow\!\texttt{CV}$\space were applied to select
the number of neighbors used for imputation, downstream linear models
obtained an external \(R^2\) of 0.7657 (0.0260) and 0.7659 (0.0259),
respectively. Similarly, downstream random forests obtained an external
\(R^2\) of 0.8097 (0.0202) and 0.8096 (0.0201), respectively. As a
reference point, the mean (standard deviation) downstream external
\(R^2\) when imputation to the mean was applied was 0.7329 (0.0283) and
0.8127 (0.0199) using linear regression and random forests,
respectively. Overall, the computational resources required to implement
$\texttt{CV}\!\circlearrowright\!\texttt{I}$\space were substantially
higher than $\texttt{I}\!\!\rightarrow\!\texttt{CV}$\space, and the
difference in downstream external \(R^2\) was \textless0.0001
(\textbf{Figure} \ref{fig:ames_cmp_time}).

\paragraph{Interpretation}

The use of $\texttt{CV}\!\circlearrowright\!\texttt{I}$\space versus
$\texttt{I}\!\!\rightarrow\!\texttt{CV}$\space resulted in a mean
relative change of 0.0262\% and -0.0220\% in downstream model
performance for linear models and random forests, respectively. These
shifts in model performance come at the cost of a \(v\)-fold increase in
the amount of computational resources allocated to handle MD. While
improvements on the order of hundredths of a percentage may be notable
for select analytic scenarios, these shifts in model performance may not
be relevant in the majority of supervised learning analyses. In the
latter case, it seems unsupervised imputation before CV can allow for
pragmatic handling of MD without sacrificing the integrity of CV.

\section{Discussion and recommendations} \label{sec:discuss}

We demonstrated empirical properties of
$\texttt{CV}\!\circlearrowright\!\texttt{I}$\space and
$\texttt{I}\!\!\rightarrow\!\texttt{CV}$\space using nearest-neighbor
imputation prior to applying regression and random forest models. We
selected these methods because they have been studied thoroughly and are
widely used in applied settings. In simulated experiments, we generated
outcomes using linear effects without interaction. We also studied three
broad scenarios that were relevant to CV: Scenario 1 was an ideal
setting where $\texttt{I}\!\!\rightarrow\!\texttt{CV}$\space and
$\texttt{CV}\!\circlearrowright\!\texttt{I}$\space should have provided
almost identical estimates of generalization error. Scenarios 2 and 3
were meant to test whether
$\texttt{I}\!\!\rightarrow\!\texttt{CV}$\space produced biased estimates
of generalization error because in settings where
$\texttt{I}\!\!\rightarrow\!\texttt{CV}$\space clearly did not mimic the
final application of a trained model to an external validation set.
Remarkably, despite its bias in scenario 2, the reduction in variance of
\(R^2\) estimates using
$\texttt{I}\!\!\rightarrow\!\texttt{CV}$\space lead to a lower overall
RMSE compared to $\texttt{CV}\!\circlearrowright\!\texttt{I}$.
Downstream model performance was consistently superior when
$\texttt{CV}\!\circlearrowright\!\texttt{I}$\space was used instead of
$\texttt{I}\!\!\rightarrow\!\texttt{CV}$. However, the increase in
performance was smaller than 1\% relative change (maximum overall
relative difference in external \(R^2\): 0.07\%). While this difference
is very small, it may be possible to find a different generative
scenario where the difference is larger. Throughout our analysis,
$\texttt{I}\!\!\rightarrow\!\texttt{CV}$\space required less computation
time than $\texttt{CV}\!\circlearrowright\!\texttt{I}$\space by a factor
of roughly \(v\), the number of folds employed by CV.

Unsupervised imputation has two interesting characteristics relevant to
predictive modeling. First, it allows for imputation of testing data
without requiring observed outcome values in those data. This
characteristic is ideal for deploying prediction models in the real
world, where outcome values are almost always unknown at the time of
prediction (otherwise, why would we be predicting them?). Second, as the
current analysis has shown, unsupervised imputation can be applied
before CV in select scenarios without inducing overly optimistic
estimates of model error. The benefits of this approach include (1)
reduced computational overhead, (2) reduced variance in model error
estimates, and (3) little difference in the performance of downstream
models. If investigators are confident that training and testing data
are identically distributed or are primarily concerned with selecting
optimal tuning parameters,
$\texttt{I}\!\!\rightarrow\!\texttt{CV}$\space may be an extremely
valuable workflow to implement. However, the drawbacks of
$\texttt{I}\!\!\rightarrow\!\texttt{CV}$\space include increased bias
for model estimation, particularly in settings similar to scenario 2
(described in Section \ref{subsec:data_gen}). If investigators are
primarily interested in estimating model error without bias and cannot
rule out the possibility that testing data are drawn from a different
population or distribution than their training data, the current study
suggests $\texttt{CV}\!\circlearrowright\!\texttt{I}$\space should be
applied instead of $\texttt{I}\!\!\rightarrow\!\texttt{CV}$. However, it
is worth noting that almost all prediction modeling decisions require
balancing bias and variance to optimize precision. Our results do not
indicate any strong difference between
$\texttt{I}\!\!\rightarrow\!\texttt{CV}$\space and
$\texttt{CV}\!\circlearrowright\!\texttt{I}$\space with regard to
precision (\textit{i.e., }RMSE, see \textbf{Table} \ref{tab:rmse}). We
suspect that in most prediction modeling applications, precision rather
than bias is of primary interest.

The current study has several strengths. We implemented computational
experiments using simulated and real data. We included different
data-generation mechanisms, different modeling procedures, different MD
patterns, and different modeling strategies to ensure our results
generalized to several common analytical settings. We examined a wide
variety of metrics to assess the benefits and weaknesses of applying
$\texttt{I}\!\!\rightarrow\!\texttt{CV}$\space versus
$\texttt{CV}\!\circlearrowright\!\texttt{I}$. We used the \texttt{ipa} R
package to conduct unsupervised imputation throughout our analyses and
disseminate both the R package and all code used for the current
analysis on the
\href{https://github.com/bcjaeger/Imputation-and-CV}{first author's
GitHub}. Each of these supplemental components ensure that our work is
easily reproduced and disseminated. There are also some gaps in the
current study that can be filled by future work. We investigated
\(v\)-fold CV in the current analysis. Future research may assess
whether these results generalize to other forms of data-splitting such
as Monte-Carlo CV or bootstrap CV. Because MNAR data present challenges
that may not be overcome by imputation alone, we did not include
simulations for MNAR data. Whether the current study's findings
generalize to settings with MNAR data remains an interesting, unanswered
question. Last, the current study has applied k-nearest neighbor
imputation throughout. As many other types of imputation procedures have
been established, there are numerous extensions of the current analysis
that may explore whether our results hold when other imputation
approaches are implemented.

\FloatBarrier

\begin{table}[p]

\caption{\label{tab:ext_rsq}True external $R^2$ mean (standard deviation) values for the modeling technique that is internally assessed using $\texttt{CV}\!\circlearrowright\!\texttt{I}$\space and $\texttt{I}\!\!\rightarrow\!\texttt{CV}$. Descriptions of scenarios 1, 2, and 3 are provided in Section \ref{subsec:data_gen}. All table values are scaled by 100 for convenience}
\centering
\begin{tabular}[t]{lcccccc}
\toprule
\multicolumn{1}{c}{ } & \multicolumn{2}{c}{Scenario 1} & \multicolumn{2}{c}{Scenario 2} & \multicolumn{2}{c}{Scenario 3} \\
\cmidrule(l{3pt}r{3pt}){2-3} \cmidrule(l{3pt}r{3pt}){4-5} \cmidrule(l{3pt}r{3pt}){6-7}
N & MAR & MCAR & MAR & MCAR & MAR & MCAR\\
\midrule
\addlinespace[0.75em]
\multicolumn{7}{l}{\textbf{10 predictors, 10 junk}}\\
\hline
\hspace{1em}100 & 37.8 (3.44) & 37.7 (3.43) & 33.5 (6.53) & 33.3 (6.69) & 33.5 (6.56) & 33.3 (6.69)\\
\hspace{1em}500 & 42.8 (2.91) & 42.7 (2.93) & 40.0 (4.97) & 39.8 (5.03) & 40.0 (4.98) & 39.8 (4.99)\\
\hspace{1em}1,000 & 43.5 (2.94) & 43.4 (2.95) & 40.9 (4.75) & 40.7 (4.80) & 40.9 (4.75) & 40.7 (4.79)\\
\hspace{1em}5,000 & 44.3 (3.01) & 44.2 (3.01) & 42.0 (4.46) & 41.8 (4.61) & 42.0 (4.46) & 41.8 (4.59)\\
\addlinespace[0.75em]
\multicolumn{7}{l}{\textbf{10 predictors, 40 junk}}\\
\hline
\hspace{1em}100 & 34.6 (3.83) & 34.5 (3.75) & 30.6 (6.39) & 30.6 (6.23) & 30.6 (6.44) & 30.5 (6.28)\\
\hspace{1em}500 & 40.6 (2.77) & 40.6 (2.77) & 38.2 (4.67) & 38.2 (4.73) & 38.2 (4.68) & 38.2 (4.61)\\
\hspace{1em}1,000 & 41.5 (2.73) & 41.5 (2.74) & 39.3 (4.68) & 39.2 (4.67) & 39.3 (4.67) & 39.2 (4.68)\\
\hspace{1em}5,000 & 42.6 (2.75) & 42.6 (2.75) & 40.6 (4.11) & 40.5 (4.17) & 40.6 (4.11) & 40.5 (4.20)\\
\addlinespace[0.75em]
\multicolumn{7}{l}{\textbf{10 predictors, 490 junk}}\\
\hline
\hspace{1em}100 & 27.5 (5.11) & 27.6 (5.03) & 23.3 (6.58) & 23.3 (6.47) & 23.3 (6.51) & 23.3 (6.47)\\
\hspace{1em}500 & 37.6 (2.93) & 37.6 (2.92) & 35.8 (4.09) & 35.9 (4.02) & 35.8 (4.09) & 35.9 (4.03)\\
\hspace{1em}1,000 & 38.8 (2.83) & 38.7 (2.83) & 37.2 (4.21) & 37.2 (4.22) & 37.2 (4.22) & 37.2 (4.23)\\
\hspace{1em}5,000 & 39.8 (2.78) & 39.8 (2.78) & 38.5 (4.06) & 38.5 (4.01) & 38.5 (4.05) & 38.5 (4.03)\\
\addlinespace[0.75em]
\multicolumn{7}{l}{\textbf{Overall}}\\
\hline
\hspace{1em}--- & 39.3 (5.51) & 39.2 (5.46) & 36.7 (7.18) & 36.6 (7.16) & 36.7 (7.17) & 36.6 (7.14)\\
\bottomrule
\end{tabular}
\end{table}

\clearpage

\begin{table}[p]

\caption{\label{tab:cv_diffs}Mean (standard deviation) absolute differences in estimates of external $R^2$ between $\texttt{CV}\!\circlearrowright\!\texttt{I}$\space and $\texttt{I}\!\!\rightarrow\!\texttt{CV}$. Descriptions of scenarios 1, 2, and 3 are provided in Section \ref{subsec:data_gen}. All table values are scaled by 100 for convenience}
\centering
\begin{tabular}[t]{lcccccc}
\toprule
\multicolumn{1}{c}{ } & \multicolumn{2}{c}{Scenario 1} & \multicolumn{2}{c}{Scenario 2} & \multicolumn{2}{c}{Scenario 3} \\
\cmidrule(l{3pt}r{3pt}){2-3} \cmidrule(l{3pt}r{3pt}){4-5} \cmidrule(l{3pt}r{3pt}){6-7}
N & MAR & MCAR & MAR & MCAR & MAR & MCAR\\
\midrule
\addlinespace[0.75em]
\multicolumn{7}{l}{\textbf{10 predictors, 10 junk}}\\
\hline
\hspace{1em}100 & 1.21 (1.02) & 1.19 (0.99) & 2.43 (1.76) & 2.44 (1.79) & 2.05 (1.64) & 2.07 (1.67)\\
\hspace{1em}500 & 0.32 (0.29) & 0.33 (0.29) & 1.34 (1.13) & 1.39 (1.19) & 0.97 (1.05) & 1.00 (1.09)\\
\hspace{1em}1,000 & 0.21 (0.20) & 0.21 (0.20) & 1.22 (1.15) & 1.28 (1.21) & 0.87 (1.07) & 0.90 (1.12)\\
\hspace{1em}5,000 & 0.09 (0.08) & 0.09 (0.09) & 1.04 (0.92) & 1.09 (1.01) & 0.69 (0.89) & 0.71 (0.95)\\
\addlinespace[0.75em]
\multicolumn{7}{l}{\textbf{10 predictors, 40 junk}}\\
\hline
\hspace{1em}100 & 1.49 (1.27) & 1.47 (1.27) & 2.71 (1.97) & 2.67 (1.91) & 2.38 (1.89) & 2.35 (1.80)\\
\hspace{1em}500 & 0.33 (0.29) & 0.34 (0.29) & 1.40 (1.23) & 1.45 (1.29) & 1.01 (1.11) & 1.04 (1.17)\\
\hspace{1em}1,000 & 0.22 (0.20) & 0.22 (0.20) & 1.27 (1.16) & 1.33 (1.23) & 0.90 (1.09) & 0.93 (1.15)\\
\hspace{1em}5,000 & 0.09 (0.09) & 0.10 (0.09) & 1.15 (1.06) & 1.23 (1.16) & 0.77 (1.01) & 0.80 (1.08)\\
\addlinespace[0.75em]
\multicolumn{7}{l}{\textbf{10 predictors, 490 junk}}\\
\hline
\hspace{1em}100 & 2.09 (1.73) & 2.02 (1.67) & 3.01 (2.18) & 3.01 (2.14) & 2.75 (2.04) & 2.79 (2.05)\\
\hspace{1em}500 & 0.34 (0.29) & 0.34 (0.29) & 1.21 (1.15) & 1.21 (1.14) & 0.88 (1.05) & 0.89 (1.05)\\
\hspace{1em}1,000 & 0.21 (0.19) & 0.21 (0.19) & 1.16 (1.16) & 1.15 (1.15) & 0.81 (1.04) & 0.79 (1.02)\\
\hspace{1em}5,000 & 0.09 (0.08) & 0.09 (0.08) & 1.12 (1.11) & 1.18 (1.17) & 0.74 (1.03) & 0.77 (1.08)\\
\addlinespace[0.75em]
\multicolumn{7}{l}{\textbf{Overall}}\\
\hline
\hspace{1em}--- & 0.56 (0.95) & 0.55 (0.93) & 1.59 (1.54) & 1.62 (1.55) & 1.23 (1.47) & 1.25 (1.48)\\
\bottomrule
\end{tabular}
\end{table}

\begin{table}[p]

\caption{\label{tab:bias}Bias of external $R^2$ estimates using $\texttt{CV}\!\circlearrowright\!\texttt{I}$\space and $\texttt{I}\!\!\rightarrow\!\texttt{CV}$. Descriptions of scenarios 1, 2, and 3 are provided in Section \ref{subsec:data_gen}. All table values are scaled by 100 for convenience}
\centering
\begin{tabular}[t]{lcccccccccccc}
\toprule
\multicolumn{1}{c}{ } & \multicolumn{6}{c}{Missing completely at random} & \multicolumn{6}{c}{Missing at random} \\
\cmidrule(l{3pt}r{3pt}){2-7} \cmidrule(l{3pt}r{3pt}){8-13}
\multicolumn{1}{c}{ } & \multicolumn{2}{c}{Scenario 1} & \multicolumn{2}{c}{Scenario 2} & \multicolumn{2}{c}{Scenario 3} & \multicolumn{2}{c}{Scenario 1} & \multicolumn{2}{c}{Scenario 2} & \multicolumn{2}{c}{Scenario 3} \\
\cmidrule(l{3pt}r{3pt}){2-3} \cmidrule(l{3pt}r{3pt}){4-5} \cmidrule(l{3pt}r{3pt}){6-7} \cmidrule(l{3pt}r{3pt}){8-9} \cmidrule(l{3pt}r{3pt}){10-11} \cmidrule(l{3pt}r{3pt}){12-13}
N & $\texttt{CV}\!\circlearrowright\!\texttt{I}$& $\texttt{I}\!\!\rightarrow\!\texttt{CV}$& $\texttt{CV}\!\circlearrowright\!\texttt{I}$& $\texttt{I}\!\!\rightarrow\!\texttt{CV}$& $\texttt{CV}\!\circlearrowright\!\texttt{I}$& $\texttt{I}\!\!\rightarrow\!\texttt{CV}$& $\texttt{CV}\!\circlearrowright\!\texttt{I}$& $\texttt{I}\!\!\rightarrow\!\texttt{CV}$& $\texttt{CV}\!\circlearrowright\!\texttt{I}$& $\texttt{I}\!\!\rightarrow\!\texttt{CV}$& $\texttt{CV}\!\circlearrowright\!\texttt{I}$& $\texttt{I}\!\!\rightarrow\!\texttt{CV}$\\
\midrule
\addlinespace[0.75em]
\multicolumn{13}{l}{\textbf{10 predictors, 10 junk}}\\
\hline
\hspace{1em}100 & 0.31 & -0.48 & 1.06 & -1.14 & 0.57 & -1.21 & 0.51 & -0.31 & 1.32 & -0.85 & 0.76 & -0.99\\
\hspace{1em}500 & 0.10 & -0.02 & 0.26 & -1.06 & -0.20 & -1.07 & 0.11 & 0.01 & 0.41 & -0.86 & -0.05 & -0.90\\
\hspace{1em}1,000 & 0.09 & 0.04 & 0.20 & -1.04 & -0.26 & -1.08 & 0.10 & 0.04 & 0.36 & -0.83 & -0.08 & -0.87\\
\hspace{1em}5,000 & 0.00 & 0.00 & 0.11 & -0.97 & -0.32 & -0.99 & 0.01 & 0.00 & 0.28 & -0.75 & -0.11 & -0.77\\
\addlinespace[0.75em]
\multicolumn{13}{l}{\textbf{10 predictors, 40 junk}}\\
\hline
\hspace{1em}100 & 0.62 & -0.52 & 1.02 & -1.44 & 0.56 & -1.56 & 0.80 & -0.39 & 1.33 & -1.17 & 0.82 & -1.31\\
\hspace{1em}500 & 0.14 & -0.01 & 0.36 & -1.01 & -0.09 & -1.02 & 0.20 & 0.07 & 0.45 & -0.88 & -0.02 & -0.91\\
\hspace{1em}1,000 & 0.04 & -0.04 & 0.19 & -1.11 & -0.29 & -1.15 & 0.06 & -0.02 & 0.22 & -1.02 & -0.22 & -1.04\\
\hspace{1em}5,000 & 0.06 & 0.03 & 0.16 & -1.06 & -0.31 & -1.09 & 0.05 & 0.03 & 0.22 & -0.92 & -0.20 & -0.94\\
\addlinespace[0.75em]
\multicolumn{13}{l}{\textbf{10 predictors, 490 junk}}\\
\hline
\hspace{1em}100 & 1.08 & -0.74 & 1.44 & -1.43 & 1.26 & -1.39 & 1.02 & -0.88 & 1.42 & -1.46 & 1.15 & -1.46\\
\hspace{1em}500 & 0.23 & 0.08 & 0.38 & -0.68 & 0.02 & -0.70 & 0.37 & 0.21 & 0.48 & -0.58 & 0.11 & -0.61\\
\hspace{1em}1,000 & 0.12 & 0.06 & 0.20 & -0.87 & -0.20 & -0.88 & 0.31 & 0.24 & 0.38 & -0.68 & -0.01 & -0.71\\
\hspace{1em}5,000 & 0.00 & -0.02 & 0.26 & -0.90 & -0.20 & -0.93 & 0.05 & 0.03 & 0.26 & -0.85 & -0.16 & -0.86\\
\addlinespace[0.75em]
\multicolumn{13}{l}{\textbf{Overall}}\\
\hline
\hspace{1em}--- & 0.23 & -0.14 & 0.47 & -1.06 & 0.04 & -1.09 & 0.30 & -0.08 & 0.59 & -0.90 & 0.16 & -0.95\\
\bottomrule
\end{tabular}
\end{table}

\begin{table}[p]

\caption{\label{tab:variance}Standard deviation of external $R^2$ estimates using $\texttt{CV}\!\circlearrowright\!\texttt{I}$\space and $\texttt{I}\!\!\rightarrow\!\texttt{CV}$. Descriptions of scenarios 1, 2, and 3 are provided in Section \ref{subsec:data_gen}. All table values are scaled by 100 for convenience}
\centering
\begin{tabular}[t]{lcccccccccccc}
\toprule
\multicolumn{1}{c}{ } & \multicolumn{6}{c}{Missing completely at random} & \multicolumn{6}{c}{Missing at random} \\
\cmidrule(l{3pt}r{3pt}){2-7} \cmidrule(l{3pt}r{3pt}){8-13}
\multicolumn{1}{c}{ } & \multicolumn{2}{c}{Scenario 1} & \multicolumn{2}{c}{Scenario 2} & \multicolumn{2}{c}{Scenario 3} & \multicolumn{2}{c}{Scenario 1} & \multicolumn{2}{c}{Scenario 2} & \multicolumn{2}{c}{Scenario 3} \\
\cmidrule(l{3pt}r{3pt}){2-3} \cmidrule(l{3pt}r{3pt}){4-5} \cmidrule(l{3pt}r{3pt}){6-7} \cmidrule(l{3pt}r{3pt}){8-9} \cmidrule(l{3pt}r{3pt}){10-11} \cmidrule(l{3pt}r{3pt}){12-13}
N & $\texttt{CV}\!\circlearrowright\!\texttt{I}$& $\texttt{I}\!\!\rightarrow\!\texttt{CV}$& $\texttt{CV}\!\circlearrowright\!\texttt{I}$& $\texttt{I}\!\!\rightarrow\!\texttt{CV}$& $\texttt{CV}\!\circlearrowright\!\texttt{I}$& $\texttt{I}\!\!\rightarrow\!\texttt{CV}$& $\texttt{CV}\!\circlearrowright\!\texttt{I}$& $\texttt{I}\!\!\rightarrow\!\texttt{CV}$& $\texttt{CV}\!\circlearrowright\!\texttt{I}$& $\texttt{I}\!\!\rightarrow\!\texttt{CV}$& $\texttt{CV}\!\circlearrowright\!\texttt{I}$& $\texttt{I}\!\!\rightarrow\!\texttt{CV}$\\
\midrule
\addlinespace[0.75em]
\multicolumn{13}{l}{\textbf{10 predictors, 10 junk}}\\
\hline
\hspace{1em}100 & 6.41 & 6.38 & 7.56 & 7.09 & 7.46 & 7.04 & 6.37 & 6.33 & 7.52 & 7.13 & 7.44 & 7.06\\
\hspace{1em}500 & 3.59 & 3.64 & 3.92 & 3.82 & 3.92 & 3.81 & 3.60 & 3.65 & 3.92 & 3.82 & 3.91 & 3.81\\
\hspace{1em}1,000 & 3.32 & 3.36 & 3.67 & 3.55 & 3.65 & 3.54 & 3.31 & 3.35 & 3.67 & 3.55 & 3.65 & 3.54\\
\hspace{1em}5,000 & 3.08 & 3.10 & 3.39 & 3.29 & 3.40 & 3.28 & 3.07 & 3.09 & 3.37 & 3.28 & 3.38 & 3.27\\
\addlinespace[0.75em]
\multicolumn{13}{l}{\textbf{10 predictors, 40 junk}}\\
\hline
\hspace{1em}100 & 6.73 & 6.62 & 7.28 & 6.91 & 7.19 & 6.88 & 6.62 & 6.51 & 7.30 & 6.89 & 7.23 & 6.85\\
\hspace{1em}500 & 3.55 & 3.60 & 3.75 & 3.65 & 3.70 & 3.65 & 3.57 & 3.61 & 3.73 & 3.64 & 3.71 & 3.65\\
\hspace{1em}1,000 & 3.14 & 3.17 & 3.31 & 3.22 & 3.32 & 3.21 & 3.16 & 3.19 & 3.28 & 3.21 & 3.31 & 3.21\\
\hspace{1em}5,000 & 2.82 & 2.84 & 3.02 & 2.89 & 3.02 & 2.89 & 2.82 & 2.84 & 2.99 & 2.88 & 3.00 & 2.88\\
\addlinespace[0.75em]
\multicolumn{13}{l}{\textbf{10 predictors, 490 junk}}\\
\hline
\hspace{1em}100 & 7.66 & 7.43 & 7.89 & 7.51 & 7.84 & 7.50 & 7.52 & 7.31 & 8.00 & 7.61 & 7.96 & 7.56\\
\hspace{1em}500 & 3.69 & 3.73 & 3.85 & 3.78 & 3.86 & 3.78 & 3.72 & 3.75 & 3.85 & 3.77 & 3.87 & 3.78\\
\hspace{1em}1,000 & 3.26 & 3.29 & 3.30 & 3.21 & 3.30 & 3.21 & 3.27 & 3.29 & 3.33 & 3.24 & 3.34 & 3.25\\
\hspace{1em}5,000 & 2.86 & 2.87 & 2.95 & 2.87 & 2.98 & 2.87 & 2.87 & 2.88 & 2.94 & 2.87 & 2.97 & 2.88\\
\addlinespace[0.75em]
\multicolumn{13}{l}{\textbf{Overall}}\\
\hline
\hspace{1em}--- & 6.50 & 6.12 & 7.31 & 6.77 & 7.31 & 6.75 & 6.52 & 6.12 & 7.34 & 6.79 & 7.33 & 6.76\\
\bottomrule
\end{tabular}
\end{table}

\begin{table}[p]

\caption{\label{tab:rmse}Root-mean-squared error of external $R^2$ estimates using $\texttt{CV}\!\circlearrowright\!\texttt{I}$\space and $\texttt{I}\!\!\rightarrow\!\texttt{CV}$. Descriptions of scenarios 1, 2, and 3 are provided in Section \ref{subsec:data_gen}. All table values are scaled by 100 for convenience}
\centering
\begin{tabular}[t]{lcccccccccccc}
\toprule
\multicolumn{1}{c}{ } & \multicolumn{6}{c}{Missing completely at random} & \multicolumn{6}{c}{Missing at random} \\
\cmidrule(l{3pt}r{3pt}){2-7} \cmidrule(l{3pt}r{3pt}){8-13}
\multicolumn{1}{c}{ } & \multicolumn{2}{c}{Scenario 1} & \multicolumn{2}{c}{Scenario 2} & \multicolumn{2}{c}{Scenario 3} & \multicolumn{2}{c}{Scenario 1} & \multicolumn{2}{c}{Scenario 2} & \multicolumn{2}{c}{Scenario 3} \\
\cmidrule(l{3pt}r{3pt}){2-3} \cmidrule(l{3pt}r{3pt}){4-5} \cmidrule(l{3pt}r{3pt}){6-7} \cmidrule(l{3pt}r{3pt}){8-9} \cmidrule(l{3pt}r{3pt}){10-11} \cmidrule(l{3pt}r{3pt}){12-13}
N & $\texttt{CV}\!\circlearrowright\!\texttt{I}$& $\texttt{I}\!\!\rightarrow\!\texttt{CV}$& $\texttt{CV}\!\circlearrowright\!\texttt{I}$& $\texttt{I}\!\!\rightarrow\!\texttt{CV}$& $\texttt{CV}\!\circlearrowright\!\texttt{I}$& $\texttt{I}\!\!\rightarrow\!\texttt{CV}$& $\texttt{CV}\!\circlearrowright\!\texttt{I}$& $\texttt{I}\!\!\rightarrow\!\texttt{CV}$& $\texttt{CV}\!\circlearrowright\!\texttt{I}$& $\texttt{I}\!\!\rightarrow\!\texttt{CV}$& $\texttt{CV}\!\circlearrowright\!\texttt{I}$& $\texttt{I}\!\!\rightarrow\!\texttt{CV}$\\
\midrule
\addlinespace[0.75em]
\multicolumn{13}{l}{\textbf{10 predictors, 10 junk}}\\
\hline
\hspace{1em}100 & 5.93 & 5.86 & 9.08 & 8.74 & 8.94 & 8.69 & 5.87 & 5.77 & 8.98 & 8.62 & 8.84 & 8.59\\
\hspace{1em}500 & 2.22 & 2.22 & 4.98 & 4.91 & 4.91 & 4.86 & 2.22 & 2.22 & 4.94 & 4.84 & 4.92 & 4.85\\
\hspace{1em}1,000 & 1.64 & 1.64 & 4.31 & 4.20 & 4.25 & 4.22 & 1.64 & 1.64 & 4.25 & 4.11 & 4.19 & 4.12\\
\hspace{1em}5,000 & 0.82 & 0.82 & 3.67 & 3.63 & 3.64 & 3.61 & 0.82 & 0.82 & 3.48 & 3.40 & 3.46 & 3.40\\
\addlinespace[0.75em]
\multicolumn{13}{l}{\textbf{10 predictors, 40 junk}}\\
\hline
\hspace{1em}100 & 6.36 & 6.26 & 8.77 & 8.53 & 8.72 & 8.59 & 6.36 & 6.23 & 8.97 & 8.64 & 8.92 & 8.68\\
\hspace{1em}500 & 2.36 & 2.36 & 4.79 & 4.69 & 4.61 & 4.59 & 2.34 & 2.33 & 4.74 & 4.61 & 4.66 & 4.62\\
\hspace{1em}1,000 & 1.66 & 1.66 & 4.39 & 4.37 & 4.39 & 4.39 & 1.69 & 1.69 & 4.42 & 4.37 & 4.39 & 4.37\\
\hspace{1em}5,000 & 0.82 & 0.82 & 3.58 & 3.48 & 3.59 & 3.52 & 0.81 & 0.81 & 3.45 & 3.35 & 3.44 & 3.36\\
\addlinespace[0.75em]
\multicolumn{13}{l}{\textbf{10 predictors, 490 junk}}\\
\hline
\hspace{1em}100 & 7.38 & 7.31 & 8.93 & 8.79 & 8.90 & 8.80 & 7.35 & 7.35 & 9.06 & 8.89 & 8.91 & 8.80\\
\hspace{1em}500 & 2.43 & 2.42 & 4.17 & 4.01 & 4.14 & 4.02 & 2.45 & 2.43 & 4.24 & 4.05 & 4.21 & 4.07\\
\hspace{1em}1,000 & 1.76 & 1.75 & 3.91 & 3.78 & 3.89 & 3.80 & 1.76 & 1.75 & 3.92 & 3.72 & 3.87 & 3.75\\
\hspace{1em}5,000 & 0.85 & 0.85 & 3.38 & 3.24 & 3.37 & 3.27 & 0.86 & 0.86 & 3.42 & 3.28 & 3.38 & 3.28\\
\addlinespace[0.75em]
\multicolumn{13}{l}{\textbf{Overall}}\\
\hline
\hspace{1em}--- & 3.62 & 3.58 & 5.74 & 5.59 & 5.68 & 5.59 & 3.61 & 3.57 & 5.75 & 5.57 & 5.68 & 5.56\\
\bottomrule
\end{tabular}
\end{table}

\begin{table}[p]

\caption{\label{tab:tune}Mean external $R^2$ when $\texttt{CV}\!\circlearrowright\!\texttt{I}$\space and $\texttt{I}\!\!\rightarrow\!\texttt{CV}$\space were applied to tune the number of neighbors used for imputation. Descriptions of scenarios 1, 2, and 3 are provided in Section \ref{subsec:data_gen}. All table values are scaled by 100 for convenience}
\centering
\begin{tabular}[t]{lcccccccccccc}
\toprule
\multicolumn{1}{c}{ } & \multicolumn{6}{c}{Missing completely at random} & \multicolumn{6}{c}{Missing at random} \\
\cmidrule(l{3pt}r{3pt}){2-7} \cmidrule(l{3pt}r{3pt}){8-13}
\multicolumn{1}{c}{ } & \multicolumn{2}{c}{Scenario 1} & \multicolumn{2}{c}{Scenario 2} & \multicolumn{2}{c}{Scenario 3} & \multicolumn{2}{c}{Scenario 1} & \multicolumn{2}{c}{Scenario 2} & \multicolumn{2}{c}{Scenario 3} \\
\cmidrule(l{3pt}r{3pt}){2-3} \cmidrule(l{3pt}r{3pt}){4-5} \cmidrule(l{3pt}r{3pt}){6-7} \cmidrule(l{3pt}r{3pt}){8-9} \cmidrule(l{3pt}r{3pt}){10-11} \cmidrule(l{3pt}r{3pt}){12-13}
N & $\texttt{CV}\!\circlearrowright\!\texttt{I}$& $\texttt{I}\!\!\rightarrow\!\texttt{CV}$& $\texttt{CV}\!\circlearrowright\!\texttt{I}$& $\texttt{I}\!\!\rightarrow\!\texttt{CV}$& $\texttt{CV}\!\circlearrowright\!\texttt{I}$& $\texttt{I}\!\!\rightarrow\!\texttt{CV}$& $\texttt{CV}\!\circlearrowright\!\texttt{I}$& $\texttt{I}\!\!\rightarrow\!\texttt{CV}$& $\texttt{CV}\!\circlearrowright\!\texttt{I}$& $\texttt{I}\!\!\rightarrow\!\texttt{CV}$& $\texttt{CV}\!\circlearrowright\!\texttt{I}$& $\texttt{I}\!\!\rightarrow\!\texttt{CV}$\\
\midrule
\addlinespace[0.75em]
\multicolumn{13}{l}{\textbf{10 predictors, 10 junk}}\\
\hline
\hspace{1em}100 & 38.1 & 38.1 & 33.9 & 34.0 & 34.0 & 34.0 & 38.3 & 38.3 & 34.2 & 34.2 & 34.1 & 34.1\\
\hspace{1em}500 & 43.8 & 43.8 & 40.7 & 40.8 & 40.8 & 40.8 & 43.9 & 43.9 & 41.0 & 41.0 & 41.0 & 41.0\\
\hspace{1em}1,000 & 44.7 & 44.8 & 41.9 & 41.9 & 41.9 & 41.9 & 44.8 & 44.8 & 42.1 & 42.1 & 42.1 & 42.1\\
\hspace{1em}5,000 & 45.7 & 45.7 & 43.3 & 43.3 & 43.3 & 43.3 & 45.8 & 45.8 & 43.5 & 43.5 & 43.5 & 43.5\\
\addlinespace[0.75em]
\multicolumn{13}{l}{\textbf{10 predictors, 40 junk}}\\
\hline
\hspace{1em}100 & 35.1 & 35.0 & 31.1 & 31.2 & 31.1 & 31.1 & 35.2 & 35.0 & 31.1 & 31.0 & 31.1 & 31.1\\
\hspace{1em}500 & 41.8 & 41.8 & 39.1 & 39.2 & 39.2 & 39.2 & 41.8 & 41.8 & 39.2 & 39.2 & 39.2 & 39.2\\
\hspace{1em}1,000 & 42.8 & 42.8 & 40.3 & 40.3 & 40.3 & 40.3 & 42.8 & 42.8 & 40.4 & 40.4 & 40.4 & 40.4\\
\hspace{1em}5,000 & 44.0 & 44.0 & 41.9 & 41.9 & 41.9 & 41.9 & 44.0 & 44.0 & 42.0 & 42.0 & 42.0 & 42.0\\
\addlinespace[0.75em]
\multicolumn{13}{l}{\textbf{10 predictors, 490 junk}}\\
\hline
\hspace{1em}100 & 28.8 & 28.5 & 24.0 & 23.7 & 24.0 & 23.7 & 28.6 & 28.4 & 24.1 & 23.8 & 24.2 & 23.9\\
\hspace{1em}500 & 39.1 & 39.1 & 37.2 & 37.3 & 37.2 & 37.3 & 39.1 & 39.1 & 37.2 & 37.3 & 37.2 & 37.2\\
\hspace{1em}1,000 & 40.4 & 40.4 & 38.7 & 38.7 & 38.7 & 38.7 & 40.4 & 40.4 & 38.7 & 38.7 & 38.7 & 38.7\\
\hspace{1em}5,000 & 41.5 & 41.5 & 40.1 & 40.1 & 40.1 & 40.1 & 41.5 & 41.5 & 40.1 & 40.1 & 40.1 & 40.1\\
\addlinespace[0.75em]
\multicolumn{13}{l}{\textbf{Overall}}\\
\hline
\hspace{1em}--- & 40.5 & 40.5 & 37.7 & 37.7 & 37.7 & 37.7 & 40.5 & 40.5 & 37.8 & 37.8 & 37.8 & 37.8\\
\bottomrule
\end{tabular}
\end{table}

\FloatBarrier

\FloatBarrier

\FloatBarrier

\FloatBarrier

\FloatBarrier

\bibliography{bibfile.bib}

\end{document}